\documentclass[11pt, a4paper]{style}
\usepackage{subcaption}
\usepackage{makecell}
\usepackage{booktabs}
\usepackage{multirow}
\usepackage{fontawesome5}
\usepackage{graphicx}
\usepackage[authoryear, sort&compress, round]{natbib}
\usepackage{amsmath,amsfonts,bm}

\def\eqref#1{equation~\ref{#1}}

\def\1{\bm{1}}

\DeclareMathAlphabet{\mathsfit}{\encodingdefault}{\sfdefault}{m}{sl}
\SetMathAlphabet{\mathsfit}{bold}{\encodingdefault}{\sfdefault}{bx}{n}

\renewcommand{\orgbranding}{%
  \includegraphics[height=30pt]{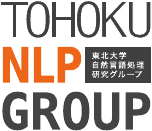}%
  \hspace{10pt}%
  \raisebox{4.5pt}{\includegraphics[height=21pt]{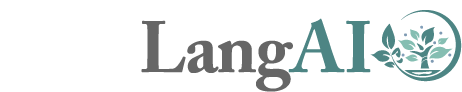}}%
}

\renewcommand{\today}{2026-06-18}

\newcommand{\titleseal}{%
  \raisebox{-0.32em}{\includegraphics[height=1.4em]{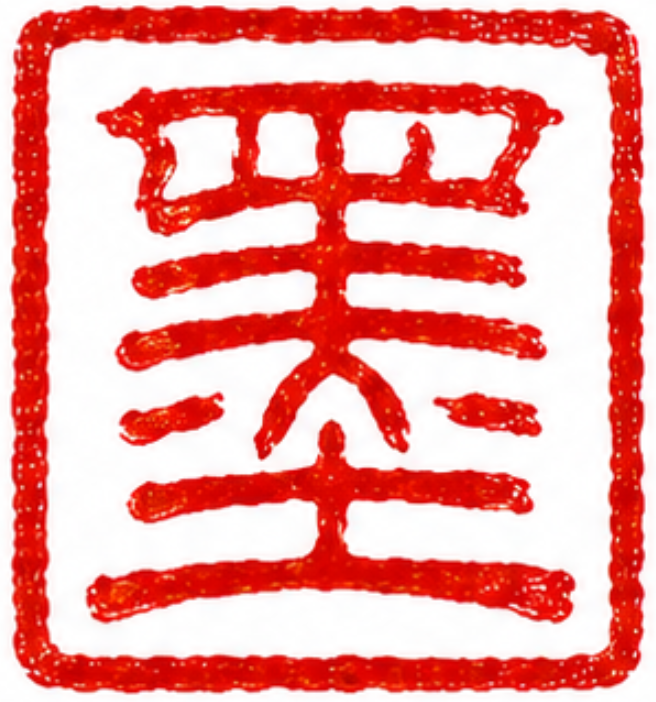}}\hspace{0.25em}}
\title{Sumi\,\titleseal: Open Uniform Diffusion \\
Language Model from Scratch}

\DeclareRobustCommand{\authorseal}[1]{%
  \,\raisebox{-0.18em}{%
    \def\rlap##1{\hbox to 0pt{##1\hss}}%
    \includegraphics[height=1.05em]{assets/#1}}}

\author[1]{Mengyu Ye\textsuperscript{*}\authorseal{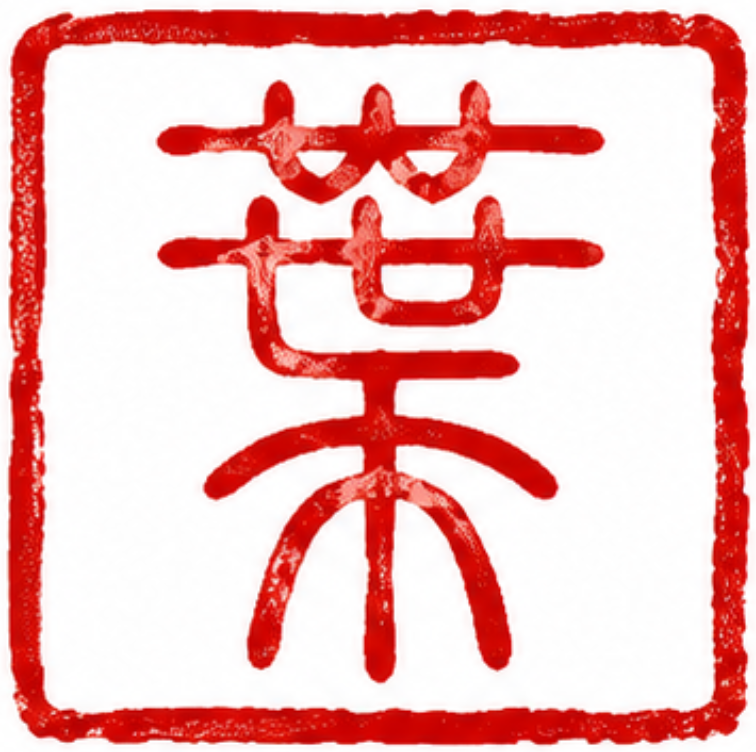}}
\author[1]{Keito Kudo\textsuperscript{*}\authorseal{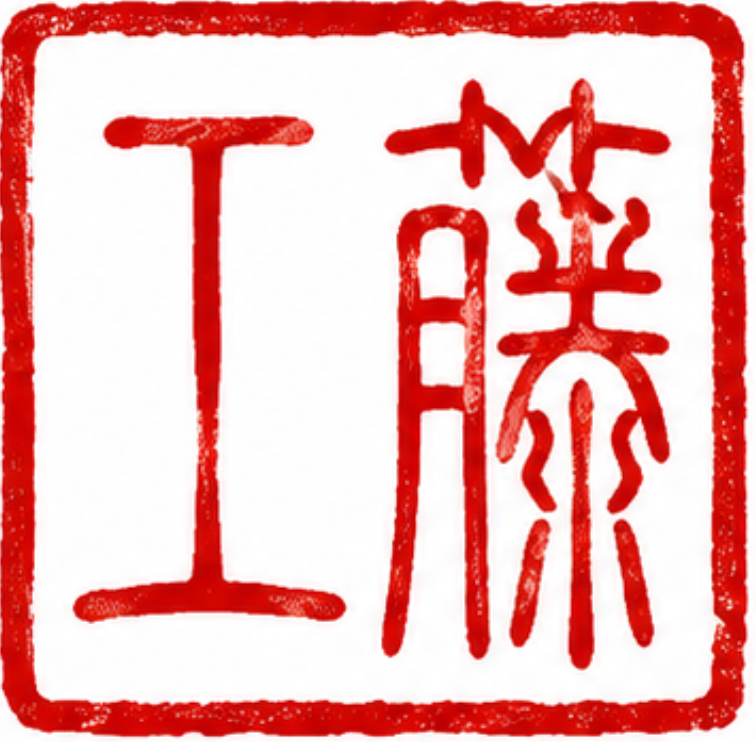}}
\author[1]{Wataru Ikeda\authorseal{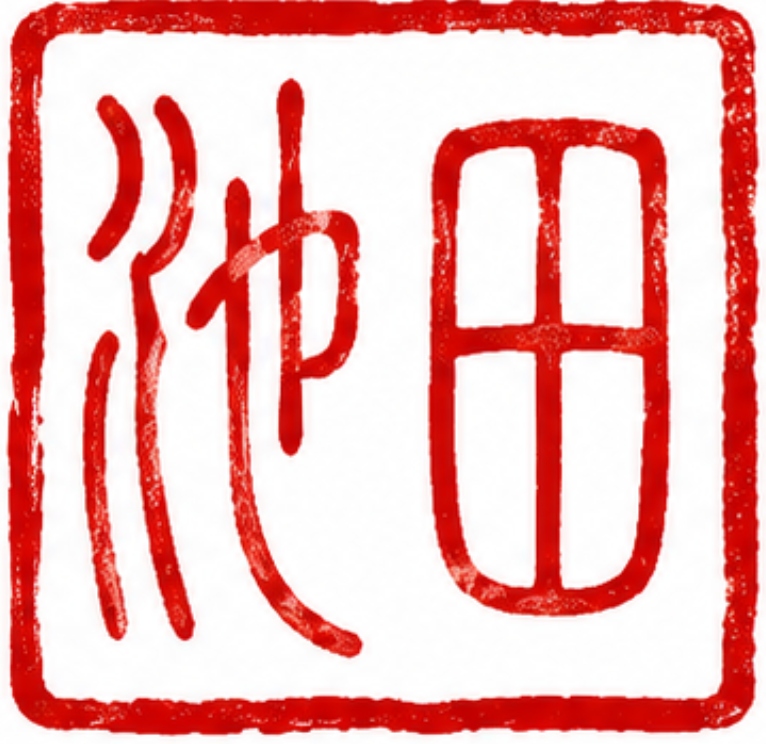}}
\author[1]{Ryosuke Matsuda\authorseal{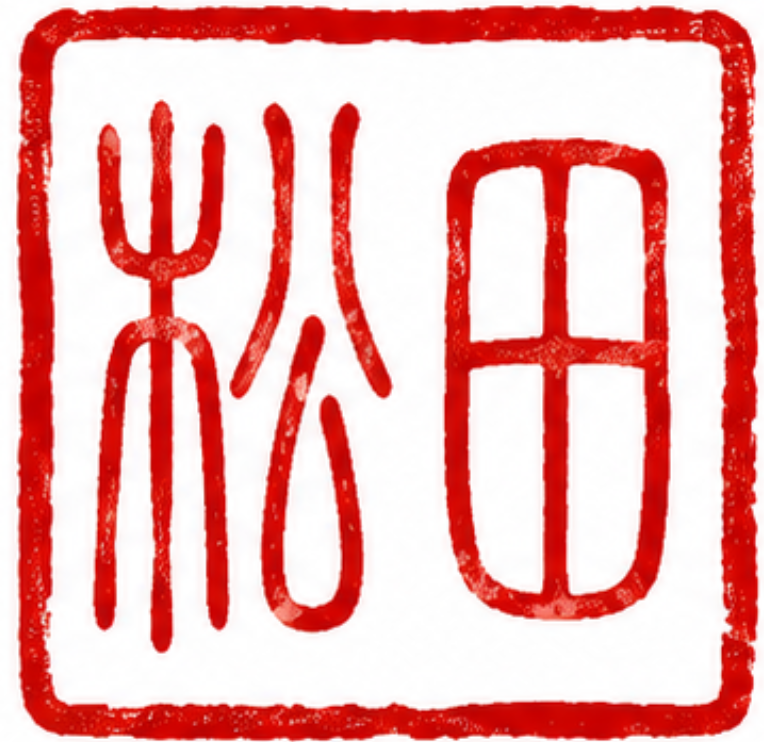}}
\author[1]{\protect\\Keisuke Sakaguchi\authorseal{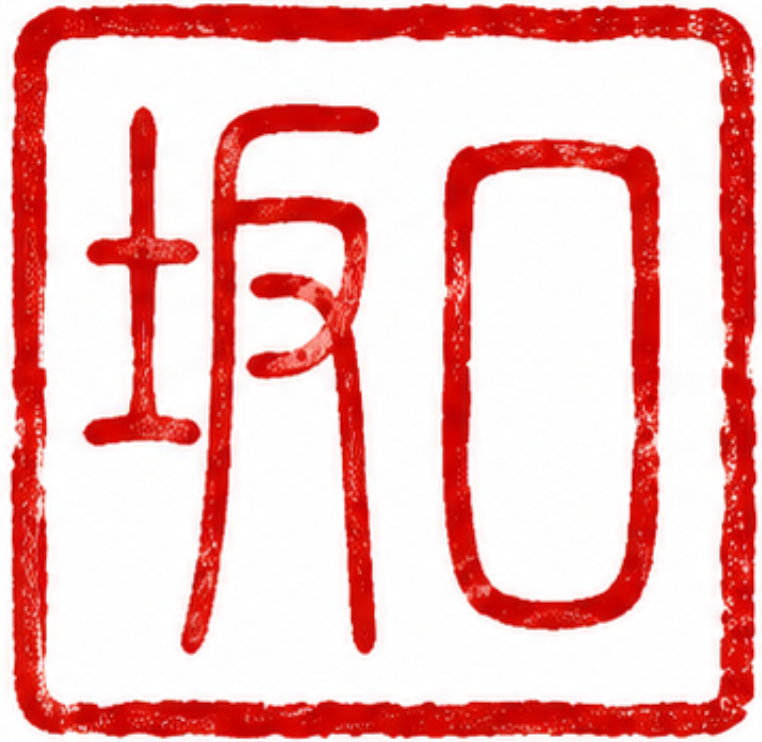}}
\author[1]{Jun Suzuki\authorseal{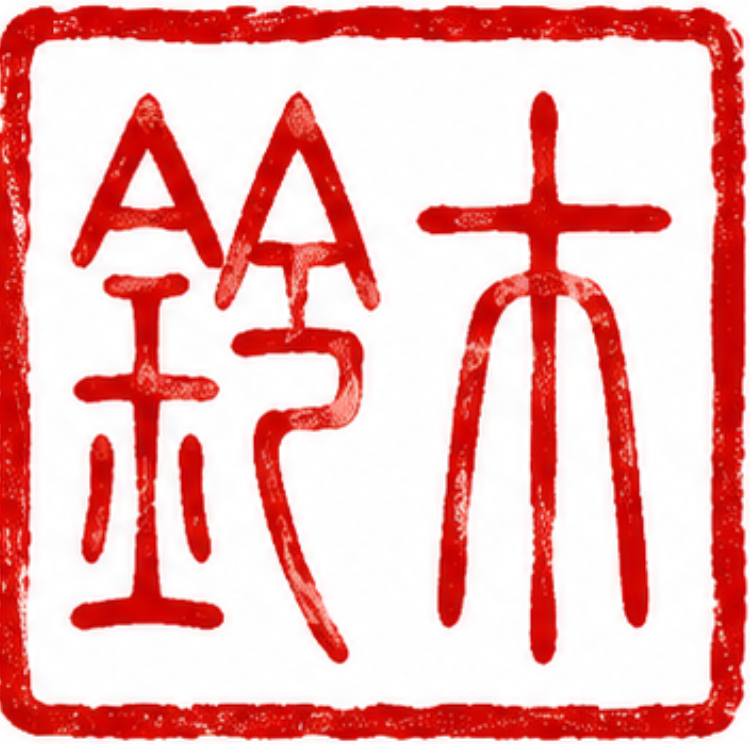}}
\affil[1]{Tohoku University}

\correspondingauthor{ye.mengyu.s1@dc.tohoku.ac.jp \& is-failab-research@grp.tohoku.ac.jp}

\equalcontribution{Equal contribution.}

\begin{abstract}
Diffusion models have become a promising alternative to autoregressive models.
Among these, uniform diffusion language models (UDLMs) permit any token to be updated at any step, in principle enabling more flexible generation.
However, no UDLM has yet been pretrained from scratch at both large parameter scale and large token budget.
Both autoregressive modeling and masked diffusion modeling already have capable models at scale that the community can study and build on; uniform diffusion has none.
A scratch-pretrained UDLM at scale would provide a clean reference point for studying scaling behavior, generation dynamics, controllability, and trade-offs against established autoregressive and masked diffusion models.
To this end, we introduce Sumi (``ink'' in Japanese), a fully open 7B uniform diffusion language model pretrained from scratch on 1.5T tokens.
Sumi performs competitively with autoregressive models trained at comparable token budgets on knowledge, reasoning, and coding benchmarks, while under-performing on commonsense benchmarks, where our education-heavy data
mixture is a likely contributor.
We release our model weights, checkpoints, and full training recipe, including a complete specification of the data mixture over publicly available corpora. 
We hope this release enables the community to study native uniform diffusion at scale and catalyzes work on its as-yet poorly understood aspects.
\end{abstract}

\begin{document}

\maketitle

\begin{center}
\href{https://www.nlp.ecei.tohoku.ac.jp/projects/sumi/}{%
  \raisebox{-0.1\height}{\faGlobe}%
  \hspace{0.4em}https://www.nlp.ecei.tohoku.ac.jp/projects/sumi/}\\[0.3em]
\href{https://huggingface.co/collections/tohoku-nlp/sumi}{%
  \raisebox{-0.22\height}{\includegraphics[height=1.1em]{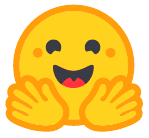}}%
  \hspace{0.4em}https://huggingface.co/collections/tohoku-nlp/sumi}\\[0.3em]
\href{https://github.com/tohoku-nlp/sumi.git}{%
  \raisebox{-0.1\height}{\faGithub}%
  \hspace{0.4em}https://github.com/tohoku-nlp/sumi}
\end{center}

\section{Introduction}

Diffusion language models have emerged as a promising alternative to autoregressive (AR) models. Masked diffusion language models (MDLMs) such as LLaDA~\citep{nie2026llada} have been scaled to 8B parameters and over 2T training tokens, reaching performance competitive with strong AR baselines. 
Uniform diffusion language models (UDLMs) relax a key rigidity of the masking process: once an MDLM fills in a masked token, that token can never be revised, whereas uniform diffusion permits any token to be updated at any step, in principle enabling more flexible generation and self-correction~\citep{rutte2025gidd}.

A recent large-scale open release, DiffusionGemma~\citep{deepmind2026diffusiongemma}, applies uniform diffusion by adapting a pretrained autoregressive model rather than pretraining from scratch.
Yet no UDLM has been pretrained from scratch at both large parameter and token scales: the largest existing models are compute-optimal checkpoints trained on comparatively small token budgets~\citep{rutte2026scaling}, and the only model trained in a data-rich regime has 1.7B parameters~\citep{sahoo2026scaling}. 
The behavior of large uniform diffusion models in the data-rich regime of modern language models therefore remains unexplored.

To this end, we introduce Sumi (``ink'' in Japanese, evoking the gradual emergence of text from noise in uniform diffusion), a fully open 7B-parameter UDLM pretrained from scratch on 1.5T tokens. 
Sumi builds on the generalized interpolating discrete diffusion~(GIDD)~\citep{rutte2025gidd} framework, together with its improved formulation~\citep{rutte2026scaling}, which reparameterizes the GIDD ELBO in terms of the signal-to-noise ratio~(SNR).
In our evaluation, Sumi performs competitively with AR models trained at comparable token budgets on general knowledge, reasoning, and coding benchmarks, while showing a noticeable gap on commonsense reasoning tasks such as PIQA, HellaSwag, and WinoGrande. Our education-heavy data mixture is a likely contributor to this gap, although we do not test this attribution directly.

Beyond the model itself, we report a small set of exploratory inference-time probes on generation tasks (\S\ref{sec:discussion}).
These run on 30 questions per task and are directional rather than conclusive, but they point to open questions we believe are worth studying in natively trained uniform diffusion models.
The most concrete observation is that Sumi generates fluently within its trained canvas range and degrades outside it, most sharply at short canvases across all four tasks and, for some tasks such as GSM8K, at long canvases as well. We use a single canvas length of 2048 throughout, which sits inside this fluent band for every task we evaluate.
The remaining observations are more tentative. Confidence-based sampling appears to impose a self-organized commitment order on an otherwise order-agnostic model; that structure permits limited parallel decoding on the coding tasks; and an explicit revision budget does not yield self-correction in our setup.
We release our model weights, checkpoints, and full training recipe, including a complete specification of the data mixture over publicly available corpora.

We hope this release enables the community to study native uniform diffusion at scale and catalyzes work on its as-yet poorly understood aspects.

\section{Training}
\subsection{Training Details}
\paragraph{Architecture.}
Sumi is a 7B-parameter time-agnostic bidirectional Transformer trained with the GIDD objective~\citep{rutte2025gidd} in its SNR-reparameterized form~\citep{rutte2026scaling}, instantiated under pure uniform noise with the log-SNR restricted to $\lambda \in [-9, 9]$ following the latter.

We use a conventional LLaMA-style block: 36 layers, hidden size 4096, SwiGLU MLPs with FFN size 12288, grouped-query attention (32 heads, 8 KV groups, head dimension 128), RMSNorm ($\epsilon = 10^{-5}$), untied input/output embeddings, no biases or dropout, and an off-by-one softmax~\citep{miller2023offbyone} in attention to mitigate attention sink~\citep{bondarenko2023quantizable, gu2025attention, xiao2024streamingllm}
We optimized for training stability. As a result, training was overall stable in both the training loss and the in-training benchmark monitor score.

We set RoPE~\citep{su2024rope} with $\theta = 500{,}000$.
We use the OLMo~3~\citep{olmo2025olmo3} tokenizer with a vocabulary size of 100{,}278, which achieves the best token efficiency on our training set.
We build our training framework based on Megatron-LM~\citep{megatron-lm}.

\paragraph{Hardware and compute.}
We train Sumi-7B on 288 NVIDIA H100 GPUs. Pre-training consumes 35{,}776 GPU-hours, and the two mid-training stages add 7{,}531 GPU-hours, for a total of 43{,}308 GPU-hours.

\paragraph{Pre-training.}
We pre-train Sumi on approximately 1.3T tokens at sequence length 1{,}184 with a global batch size of 4{,}608 sequences ($\approx$5.5M tokens per step), in bfloat16; the sequence length was chosen to maximize throughput on our hardware. 
Following \citet{rutte2026scaling}, we minimize the unweighted ELBO as a surrogate loss while reporting the true ELBO for evaluation.
We use AdamW~\citep{loshchilov2018decoupled} with $\beta = (0.9, 0.95)$, weight decay 0.1, gradient clipping at 1.0, and an auxiliary z-loss~\citep{3648699.3648939} with coefficient $10^{-5}$.

\paragraph{Mid-training.}
After pre-training, we perform two mid-training stages on a domain-specific data mixture (\S~\ref{sec:train-data}): 130B tokens at the original sequence length of 1{,}184, followed by 120B tokens at an extended length of 4{,}864, with 1{,}152 sequences per batch to keep the token batch size approximately constant at $\approx$5.6M tokens per step.

A single WSD~\citep{hu2024minicpm} learning-rate schedule spans all three stages: a 2{,}000-step warmup to a peak of $2 \times 10^{-4}$, a constant phase, and a 2{,}000-step linear cooldown to $2 \times 10^{-5}$ at the end of the final mid-training stage. 

\begin{figure}[t]
\centering
\begin{subfigure}{0.48\linewidth}
  \includegraphics[width=\linewidth]{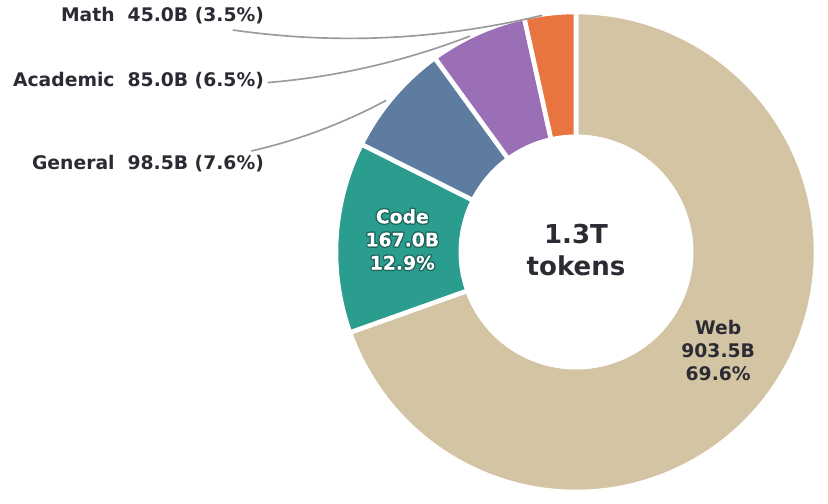}
  \caption{Pre-training (1.3T tokens).}
\end{subfigure}
\hfill
\begin{subfigure}{0.48\linewidth}
  \includegraphics[width=\linewidth]{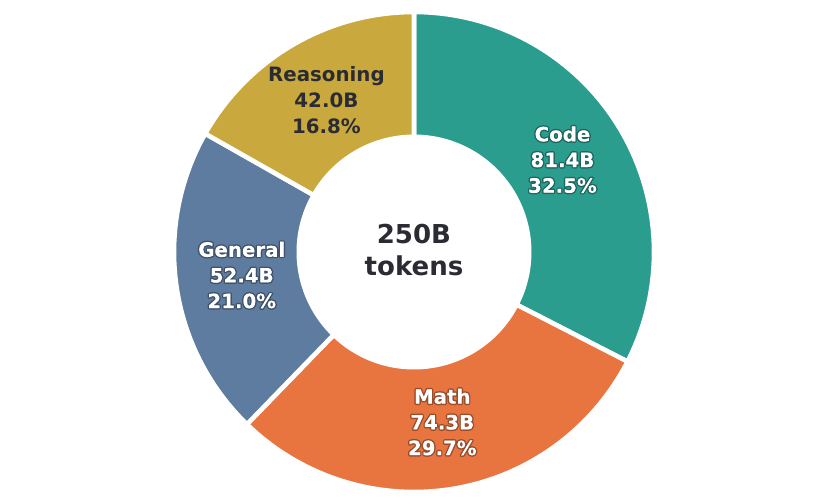}
  \caption{Mid-training (250B tokens).}
\end{subfigure}
\caption{Training data composition. Pre-training web data is filtered and re-ranked by educational score, and the mid-training mixture is weighted toward code, math, and reasoning over general text. This composition trades coverage of casual web text for knowledge- and reasoning-oriented content.}
\label{fig:data-mix}
\end{figure}

\subsection{Training Data}
\label{sec:train-data}
\paragraph{Pre-training.}
Our pre-training corpus is built on llm-jp-corpus-v4 \citep{llmjp-corpus-v4}.
We use only the English portion of the corpus and exclude its FineWeb-derived subsets, retaining the remaining English subsets together with the StarCoder code subset (\texttt{code\_olmo-starcoder}).
We additionally include Python code from swallow\_code\_v2 \citep{fujii2025swallowcode}, a dataset that is also incorporated into the llm-jp-v4 corpus.
The remainder of the budget is filled with the en\_fineweb-rescored subset of llm-jp-corpus-v4.1 \citep{llmjp-corpus-v4.1}, in which each FineWeb document is assigned an educational score following the FineWeb-Edu methodology \citep{lozhkov2024fineweb-edu}, using a lightweight classifier distilled from Qwen3-32B~\citep{yang2025qwen3} annotations; we follow the LLM-jp-4 data recipe in using this rescored version in place of the original FineWeb split.
We selected the subsets in descending order of their scores to construct a 1.3T-token dataset.
Figure~\ref{fig:data-mix}a summarizes the resulting composition.

\paragraph{Mid-training.}
We base the mid-training mixture on the English portion of llm-jp-corpus-midtraining-v2 \citep{kodama2026llmjpmid}, used as released except for three changes.
First, we downsample the \texttt{nemotron\_\allowbreak pretraining\_\allowbreak code\_\allowbreak v2} from its full 491B tokens to 63B to avoid over-weighting code.
Second, we additionally include en\_megamath-web-pro-max-oss from llm-jp-corpus-v4.1, an open reconstruction of MegaMath-Web-Pro-Max \citep{wang2025megamath} in which gpt-oss-120b~\citep{openai2025gptoss} is used to annotate and paraphrase documents; we filter it by the \texttt{math\_score} field inherited from the original MegaMath corpus \citep{zhou2025megamath}, retaining the highest-scoring documents up to 48.9B tokens.
Third, for the reasoning data we retain only samples of up to approximately 4{,}096 tokens, matching the model's context length, which keeps about 45\% of the bucket's tokens.
The resulting mixture totals approximately 250B tokens across four buckets (Figure~\ref{fig:data-mix}b): coding (81.4B, 32.5\%), math (74.3B, 29.7\%), general (52.4B, 21.0\%), and reasoning (42.0B, 16.8\%).
\footnote{All token counts are measured with OLMo~3 tokenizer we use and therefore differ from those reported in the original corpus releases.}

\paragraph{Data availability.}
All training data is drawn from publicly available corpora, available from their respective sources as referenced above; rather than redistributing them, we document the selection procedure in this section and the resulting mixture in Figure~\ref{fig:data-mix}.
Together with the upstream releases, we believe this is sufficient to reconstruct a functionally equivalent training corpus.

\section{Evaluation}
\subsection{Evaluation Settings}
We run all evaluations with the lm-evaluation-harness~\citep{gao2024evalharness}, modified only to support diffusion-based scoring.
We distinguish two quantities throughout our evaluation. The \emph{canvas length} is the number of token positions allocated at the start of generation, i.e., the size of the buffer the model fills. The \emph{generation length} is the number of positions actually scored. For likelihood tasks, positions beyond the answer and up to the generation length are filled with random tokens; for generation tasks, token updates are permitted only within the generation length, while positions beyond it are held at their random initialization.

We insert an \texttt{<EOS><BOS>} boundary between the generation region and the trailing random tokens, matching the packing distribution seen during training. This is a workaround: we did not apply an attention mask during training, because we optimized for throughput and the fastest kernel available on our hardware does not support custom attention masking.

We use a canvas length of 2048 for all tasks. We find that the model performs best at this length and degrades on substantially shorter or longer canvases. This matches the default behavior of our released generation function, which initializes 2048 token positions, places the prompt and the requested generation length, separates the two with the \texttt{<EOS><BOS>} boundary, and fills the remainder with random tokens, following \citep{rutte2026scaling}.

For the generation tasks, we set the generation length to 512 for BBH, 64 for GSM8K, and 256 for both HumanEval and MBPP. For likelihood tasks, we fill the remainder of the canvas with random tokens.

\begin{table}[t]
\centering
\caption{Benchmark result of Sumi-7B. Values are accuracy/score, with the number of in-context shots in parentheses. Bold marks the best score among models evaluated under our protocol. $^{*}$\,marks generation-based tasks; all other benchmarks are scored by a Monte Carlo estimate of the conditional likelihood (ELBO). $^{\dagger}$\,and $^{\ddagger}$\,denote scores taken from the LLaDA paper~\citep{nie2026llada} and the Dream paper~\citep{ye2025dream}, respectively.}
\label{tab:main-results}
\setlength{\tabcolsep}{5pt}
\resizebox{\textwidth}{!}{%

\begin{tabular}{l cccc|cc}
\toprule
& \multicolumn{4}{c}{\textbf{Evaluated under our protocol}}
  & \multicolumn{2}{c}{\textbf{Reported by prior work}} \\
\cmidrule(lr){2-5} \cmidrule(lr){6-7}
 & \textbf{Sumi-7B} & \textbf{Falcon-7B} & \textbf{Llama\,2-7B} & \textbf{OLMo-7B} & \textbf{LLaDA-8B} & \textbf{Llama\,3-8B} \\
\cmidrule(lr){2-2}\cmidrule(lr){3-3}\cmidrule(lr){4-4}\cmidrule(lr){5-5}\cmidrule(lr){6-6}\cmidrule(lr){7-7}
Paradigm        & Uniform Diffusion & AR     & AR   & AR     & Masked Diffusion & AR \\
Training Tokens & $1.5$T            & $1.5$T & $2$T & $2.5$T & $2.3$T & $15$T \\
Training Data   & Fully Released        & Partially Released & Not Released & Fully Released & Not Released & Not Released \\
\midrule
 & \multicolumn{6}{c}{General Knowledge} \\
\cmidrule(lr){2-7}
MMLU       & $\mathbf{51.1}\,(5)$ & $27.2\,(5)$ & $46.0\,(5)$ & $28.0\,(5)$ & $65.9^{\dagger}\,(5)$ & $65.4^{\dagger}\,(5)$ \\
RACE       & $\mathbf{41.4}\,(0)$ & $38.3\,(0)$ & $39.5\,(0)$ & $37.9\,(0)$ & $38.7^{\ddagger}\,(0)$ & $39.2^{\ddagger}\,(0)$ \\
TruthfulQA & $\mathbf{46.6}\,(0)$ & $34.3\,(0)$ & $38.8\,(0)$ & $35.9\,(0)$ & $46.1^{\dagger}\,(0)$ & $44.0^{\dagger}\,(0)$ \\
\midrule
 & \multicolumn{6}{c}{Reasoning \& Math} \\
\cmidrule(lr){2-7}
GSM8K$^{*}$   & $\mathbf{32.8}\,(4)$ & $5.3\,(4)$  & $13.5\,(4)$ & $3.8\,(4)$  & $70.3^{\dagger}\,(4)$ & $48.7^{\dagger}\,(4)$ \\
ARC-Easy      & $70.0\,(0)$ & $70.8\,(0)$ & $\mathbf{73.8}\,(0)$ & $68.8\,(0)$ & $71.8^{\ddagger}\,(0)$ & $81.1^{\ddagger}\,(0)$ \\
ARC-Challenge & $43.0\,(0)$ & $43.2\,(0)$ & $\mathbf{45.1}\,(0)$ & $40.3\,(0)$ & $45.9^{\dagger}\,(0)$ & $53.1^{\dagger}\,(0)$ \\
BBH$^{*}$     & $31.8\,(3)$ & $27.1\,(3)$ & $\mathbf{39.6}\,(3)$ & $29.8\,(3)$ & $49.7^{\dagger}\,(3)$ & $62.1^{\dagger}\,(3)$ \\
GPQA          & $\mathbf{26.1}\,(5)$ & $24.6\,(5)$ & $24.3\,(5)$ & $24.8\,(5)$ & $25.2^{\dagger}\,(5)$ & $25.9^{\dagger}\,(5)$ \\
\midrule
 & \multicolumn{6}{c}{Coding} \\
\cmidrule(lr){2-7}
HumanEval$^{*}$ & $\mathbf{22.6}\,(0)$ & $0.0\,(0)$  & $12.8\,(0)$ & $13.4\,(0)$ & $35.4^{\dagger}\,(0)$ & $34.8^{\dagger}\,(0)$ \\
MBPP$^{*}$      & $\mathbf{26.6}\,(3)$ & $12.4\,(3)$ & $23.2\,(3)$ & $21.4\,(3)$ & $40.0^{\dagger}\,(4)$ & $48.8^{\dagger}\,(4)$ \\
\midrule
 & \multicolumn{6}{c}{Commonsense} \\
\cmidrule(lr){2-7}
PIQA       & $66.4\,(0)$ & $\mathbf{80.5}\,(0)$ & $78.7\,(0)$ & $79.8\,(0)$ & $73.6^{\dagger}\,(0)$ & $80.6^{\dagger}\,(0)$ \\
HellaSwag  & $60.0\,(0)$ & $\mathbf{76.3}\,(0)$ & $76.2\,(0)$ & $75.6\,(0)$ & $70.5^{\dagger}\,(0)$ & $79.1^{\dagger}\,(0)$ \\
WinoGrande & $60.0\,(5)$ & $71.6\,(5)$ & $\mathbf{74.7}\,(5)$ & $71.3\,(5)$ & $74.8^{\dagger}\,(5)$ & $77.3^{\dagger}\,(5)$ \\
\bottomrule
\end{tabular}
}
\end{table}

\subsection{Benchmarks}
We evaluate Sumi on 13 benchmarks across four categories:

\textbf{General knowledge:} MMLU~\citep{hendrycks2021mmlu}, RACE~\citep{lai2017race}, and TruthfulQA~\citep{lin2022truthfulqa}.

\textbf{Reasoning and mathematics:} ARC-Easy and ARC-Challenge~\citep{clark2018arc}, GPQA~\citep{rein2024gpqa}, BIG-Bench Hard~\citep{suzgun2023bbh}, and GSM8K~\citep{cobbe2021gsm8k}. 

\textbf{Coding:} HumanEval~\citep{chen2021humaneval} and MBPP~\citep{austin2021mbpp}.

\textbf{Commonsense:} WinoGrande~\citep{sakaguchi2020winogrande}, PIQA~\citep{bisk2020piqa}, and HellaSwag~\citep{zellers2019hellaswag}.

\subsection{Baseline Models}

Sumi is trained on 1.5T tokens. To maximize comparability, we evaluate three open autoregressive models of similar parameter count and comparable token budget under the same protocol as Sumi: Falcon-7B~\citep{almazrouei2023falcons}, Llama~2-7B~\citep{touvron2023llama2}, and OLMo-7B~\citep{groeneveld2024olmo}.
We additionally report reference scores for LLaDA-8B and Llama~3-8B, taken from \citet{nie2026llada} and \citet{ye2025dream}; these are not evaluated under our protocol and are provided for context only.
Among models above, Sumi and OLMo are the only two models that fully released its training data.

\subsection{Results}
Table~\ref{tab:main-results} reports the benchmark scores. On general knowledge and coding, Sumi achieves the best scores among the models evaluated under our protocol, reflecting the educational and code-heavy composition of our data mixture. 
On reasoning and mathematics, where the relevant data is comparatively limited in our mid-training mixture, Sumi is competitive with Llama~2-7B and mostly ahead of Falcon-7B. 

On commonsense, Sumi is among the weakest of the models we evaluate.
Our education-heavy data mixture is a likely contributor to this gap. 
Educational and quality filtering has been observed to improve knowledge- and reasoning-intensive benchmarks while degrading commonsense benchmarks such as HellaSwag and PIQA~\citep{penedo2024fineweb, allal2025smollm2}.
We caution against reading the data mixture as a complete explanation: the gap is too large to attribute to data composition alone, and we leave a fuller account to future work.

\begin{figure}[t]
\centering
\includegraphics[width=\linewidth]{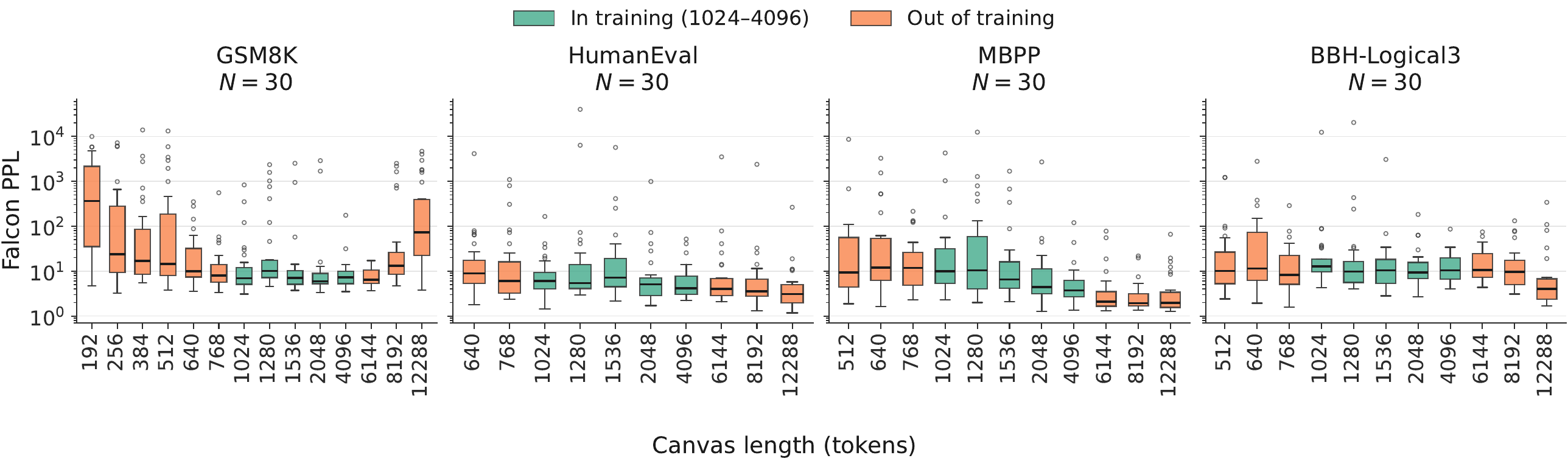}
\caption{Generation fluency as a function of canvas length, measured as Falcon-7B perplexity over Sumi's generations on 30 sampled questions per task (lower is better). Sumi is trained at sequence lengths 1184 (pre-training and mid-training) and 4864 (length extension). Green marks the swept canvas lengths that fall within this trained range; orange marks lengths below or above it.}
\label{fig:canvas-ppl}
\end{figure}

\section{Discussion}
\label{sec:discussion}
We close with a set of small, directional analyses of Sumi's generation behavior. These are exploratory probes rather than controlled claims, and we report them to motivate future study of natively trained uniform diffusion models. 

All experiments use 30 questions sampled uniformly at random from each of four generation tasks: GSM8K, HumanEval, MBPP, and the BBH subtask \texttt{logical\_deduction\_three\_objects} (BBH-Logic3 for brevity). 
We choose this specific BBH subtask because its answer format extracts cleanly and because Sumi attains its strongest BBH score on it (74.8). 
We probe four questions. The first identifies the canvas-length band within which Sumi generates fluently at all; this is the regime in which the remaining scopes operate. Those concern how the model commits tokens within that band: under which sampler, in what order, how many per step, and whether reversibly.

In summary, our early evidence suggests that Sumi is fluent only within a canvas-length band whose width is task-dependent, though a single setting serves all our tasks well; that within this band, the confidence-based adaptive sampler imposes useful, task-shaped structure on an otherwise order-free generation process; that this structure buys limited parallelism for free on the coding tasks; and that naive extra compute does not translate into self-correction.
We stress again that all discussions here are directional observations on small samples, intended to point toward questions worth studying in natively trained uniform diffusion models rather than to settle them in this report.

\begin{figure}[t]
\centering
\includegraphics[width=\linewidth]{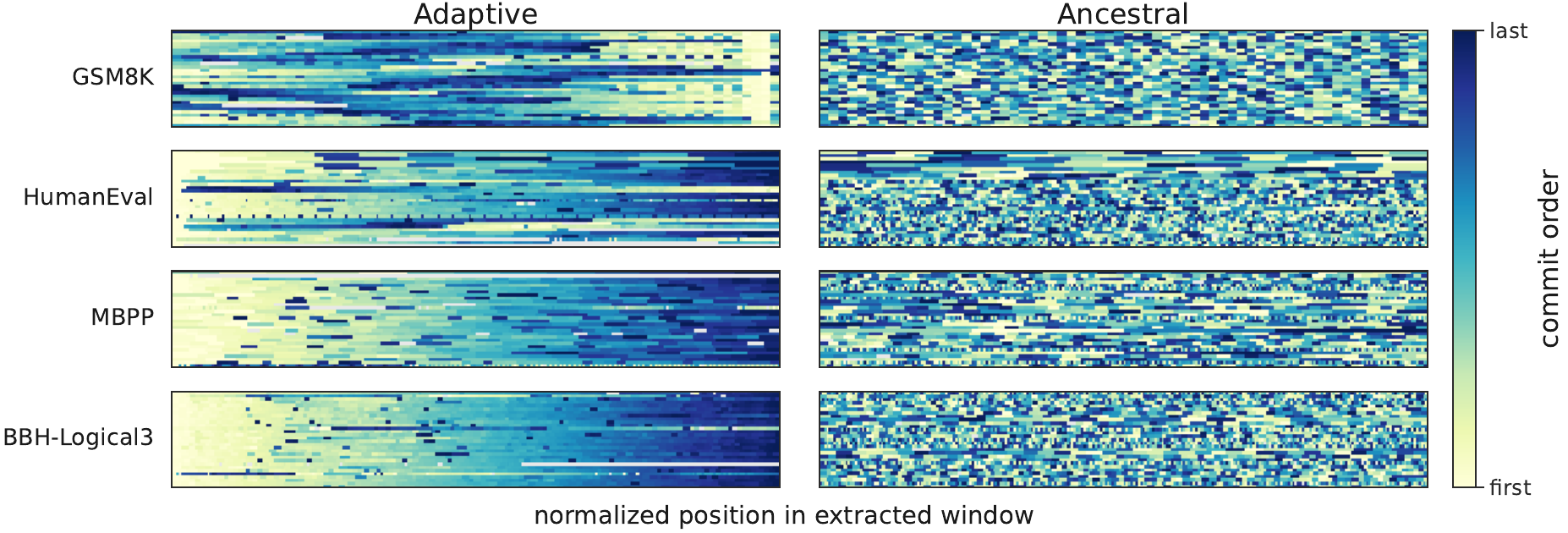}
\caption{Per-position commit order in the extracted answer window under adaptive (confidence) sampling versus ancestral sampling, for 30 sampled questions per task. Each row is one generation; color encodes the first denoising step at which a position reaches its final token (light: early, dark: late).}
\label{fig:commit-order}
\end{figure}

\subsection{Is the usable generation canvas task-dependent?}
We sweep the generation canvas length and measure fluency as the perplexity of Falcon-7B over Sumi's generations, holding the per-task generation length fixed to the setting used in Table~\ref{tab:main-results}. Falcon perplexity is only a rough fluency proxy, but its trend across canvas lengths is informative. 

Figure~\ref{fig:canvas-ppl} shows that fluency depends on both canvas length and task, with the task seems to be the dominant factor. GSM8K is the most fragile: once the canvas leaves the training range its perplexity explodes and the model emits near-random text, most severely at short canvases. The other three tasks degrade far more gently, and their degradation is concentrated at short canvases; 
at long canvases, out to roughly 2.5 times the longest trained sequence length, perplexity stays flat or even decreases.

\subsection{Does confidence sampling impose an order on token commitment?}
Confidence sampling~\citep{rutte2026scaling} improves Sumi's scores substantially in generation-based tasks.
The sampler is order-agnostic by construction: it selects positions that remain noisy under $p_{\mathrm{prior}}(z_t)$ and where the model assigns high probability to some token other than the current one, i.e., where the potential improvement
$\max_{z'} p_\theta(x = z' \mid z_t) - p_\theta(x = z_t \mid z_t)$ is large. 
We hypothesis the gains could be explained by the model committing the positions it can already determine and deferring the rest, an adaptive schedule that the sampler discovers rather than encodes. 
To probe this, we log each position's commit order, defined as the first denoising step at which it reaches its final value, restricting the analysis to the extracted answer window.

Figure~\ref{fig:commit-order} contrasts adaptive with ancestral sampling. Under adaptive sampling the per-position commit order is self-organized and visible across generations, whereas under ancestral sampling it is essentially unstructured. A model trained with a fully order-agnostic objective therefore does not commit in a fixed canonical order by default; confidence guidance is what induces the structure.

\subsection{How does the parallel decoding affect the generation quality?}
Masked diffusion LMs are reported to need one token per step for best accuracy, which limits parallel decoding in practice. We test whether the same holds for Sumi by committing $k$ tokens per step for $k \in \{1, 2, 4, 8, 16, 32\}$ and counting correct answers. 

\begin{figure}[t]
\centering
\includegraphics[width=\linewidth]{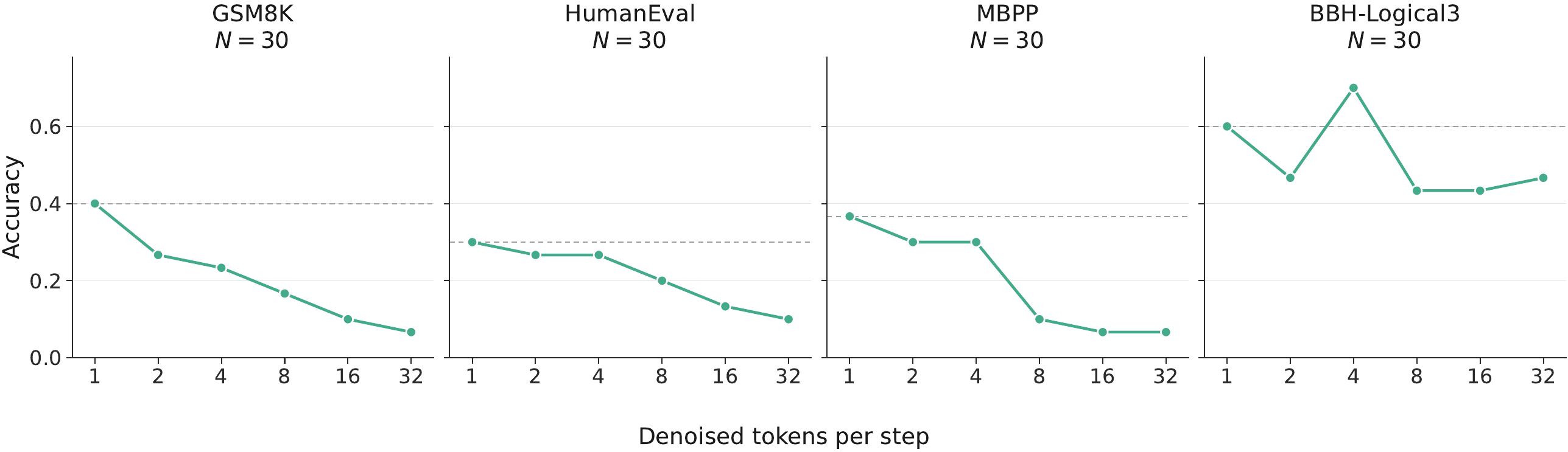}
\caption{Accuracy when committing $k$ tokens per denoising step ($k \in \{1, 2, 4, 8, 16, 32\}$) on 30 sampled questions per task. The dashed line marks the single-token ($k=1$) per step baseline.}
\label{fig:parallel-acc}
\end{figure}

Figure~\ref{fig:parallel-acc} shows that, outside GSM8K, accuracy is largely preserved up to four tokens per step: HumanEval and MBPP stay within one to two samples of the single-token baseline through $k = 4$, with MBPP dropping sharply only at $k = 8$. GSM8K is the exception and degrades immediately, already losing accuracy at $k = 2$. BBH-Logic3 is non-monotonic, and with only 30 samples its apparent peak at $k = 4$ is more like sampling noise rather than a reliable gain. 
The broad picture is that uniform diffusion admits modest parallel decoding on the coding tasks, while multi-step arithmetic remains order-sensitive; the logical-deduction subtask is inconclusive at our sample size.

\subsection{Given a revision budget, does the model correct its own tokens?}
Uniform diffusion can in principle overwrite committed tokens, so we ask whether extra denoising yields self-correction. We over-denoise the generation region, running it for one, two, four, and eight times the generation length, which gives the model 0, 1, 3, and 7 additional passes to revise already-committed tokens (the revision budget), to probe whether the model really improve the already denoised tokens.

Table~\ref{tab:revision} reports the result. A large fraction of revision steps do overwrite a committed token (58\% to 100\%), yet the net effect is negligible: at most 1\% of final tokens differ from the first pass, extracted answers almost never flip (at most one in 30), and accuracy is unchanged on every task. Inspecting the trajectories, the overwrites are predominantly $A \to B \to A$ round trips rather than directed edits. One reading is that Sumi's capability on these tasks is near a ceiling and the model does not know how to improve a committed answer; whether a revision setup designed to target errors would change this is left to future work.

\begin{table}[t]
\centering
\caption{Effect of an explicit revision budget on Sumi's generations. \emph{Edits} is the fraction of revision steps that overwrite at least one committed position; \emph{net token change} is the fraction of final tokens differing from the first-pass commit; \emph{flips} counts extracted answers that change.}
\label{tab:revision}
\begin{tabular}{lccc}
\toprule
Task & Edit steps (\%) & Net token change (\%) & Answers changed \\
\cmidrule(lr){1-1} \cmidrule(lr){2-2} \cmidrule(lr){3-3} \cmidrule(lr){4-4}
GSM8K      & 58  & 1.0 & 1/30  \\
HumanEval  & 89  & 0.2 & 0/30  \\
MBPP       & 100 & 0.4 & 0/30  \\
BBH-Logic3 & 100 & 0.1 & 0/30  \\
\bottomrule
\end{tabular}
\end{table}

\section{Conclusion and Future Work}
\label{sec:conclusion}
We introduce Sumi, a fully open 7B uniform diffusion language model (UDLM)
pretrained from scratch on 1.5T tokens, the first UDLM natively trained at both
large parameter and token scale.
Sumi performs competitively with autoregressive models trained at comparable
token budgets on general knowledge, reasoning, and coding benchmarks, while
underperforming on commonsense benchmarks; our education- and code-heavy data
mixture is a likely contributor to this gap.
We release our model weights, intermediate checkpoints, and full training recipe,
including a complete specification of the data mixture over publicly available
corpora, so that the community can study uniform diffusion at scale.

Beyond the model, our inference-time probes surface four observations about
natively trained uniform diffusion.
Generation fluency is sensitive to canvas length within a task-dependent band; a
single canvas length of 2048 sits inside this band for all our tasks and is what
we use throughout evaluation.
Within that band, confidence sampling induces a self-organized commitment order
on an otherwise order-agnostic model.
This structure admits modest parallel decoding on the coding tasks, while
multi-step arithmetic remains order-sensitive. Finally, an explicit revision
budget does not yield self-correction: extra denoising overwrites committed
tokens but the edits are predominantly $A \to B \to A$ round trips that leave the
final output and accuracy unchanged.
We read these probes as directional rather than conclusive.

We see several directions for future work. We are preparing an instruction-tuned
variant of Sumi and will release it in a future update.
The absence of self-correction under a generic revision budget leaves open
whether a setup designed to target likely errors would recover it.
More broadly, controlled comparisons under a matched evaluation protocol would
clarify which of Sumi's generation behaviors are intrinsic to uniform diffusion,
a question this report raises but does not settle.

\section*{Limitations \& Risks}
\label{sec:limitations}
Our inference-time analyses are directional rather than conclusive. They run on
30 sampled questions per task and use Falcon-7B perplexity as a rough fluency
proxy, and we do not run the matched comparisons against masked diffusion and
autoregressive models that would be needed to attribute these behaviors to the
uniform diffusion paradigm itself rather than to Sumi specifically.
The commonsense gap is larger than our education- and code-heavy data mixture
can account for on its own, and we do not test this data-composition attribution
directly.

Sumi is released as a pretrained base model and has undergone no instruction
tuning, alignment, or safety filtering. It therefore inherits the risks common
to such models.
Adversarial or careless prompting can elicit harmful, offensive, or otherwise
sensitive text, and comparable outputs can arise unprompted, for example as a
reflection of biases in the pretraining corpus.
Sumi likewise has no mechanism for ensuring factual accuracy and may state false
information with apparent confidence. We release it to support research rather
than direct deployment, and we encourage anyone building on it to weigh these
risks for their own setting and to verify factual claims independently.

\section*{Acknowledgment}
This work was carried out using the TSUBAME4.0 supercomputer at Institute of Science Tokyo, provided through the TSUBAME Grand Challenge Large-Scale Computing Program, whose generous computational resources we gratefully acknowledge.
For the evaluation experiments, we used computational resources offered under the category of HPCI Research Projects by the Research Institute for Information Technology, Kyushu University, and the ABCI 3.0 system provided by AIST and AIST Solutions with support from ``ABCI 3.0 Development Acceleration Use.''
This work was supported by the ``R\&D Hub Aimed at Ensuring Transparency and Reliability of Generative AI Models'' project of the Ministry of Education, Culture, Sports, Science and Technology; JST Moonshot R\&D Grant Number JPMJMS2011-35 (fundamental research); JSPS KAKENHI Grant Numbers JP25KJ0615; JST BOOST, Japan Grant Number JPMJBS2421.

\bibliography{references}

\clearpage

\end{document}